\documentclass{article}

\usepackage{arxiv}
\usepackage{caption}
\usepackage{subfig}

\usepackage[utf8]{inputenc} 
\usepackage[T1]{fontenc}    
\usepackage{hyperref}       
\usepackage{url}            
\usepackage{booktabs}       
\usepackage{amsfonts}       
\usepackage{nicefrac}       
\usepackage{microtype}      
\usepackage{lipsum}		
\usepackage{graphicx}
\usepackage{natbib}
\usepackage{doi}

\title{Exploring the efficacy of neural networks for trajectory compression and the inverse problem}

\author{ Theodoros Ntakouris \\
	Dept. of Computer Engineering and Informatics \\
	University of Patras \\
	\texttt{research@ntakour.is} \\
}

\renewcommand{\headeright}{}
\renewcommand{\undertitle}{Technical Report}
\renewcommand{\shorttitle}{Technical Report}

\hypersetup{
pdftitle={Exploring the effifacy of neural networks for trajectory compression and inverse lookup},
pdfsubject={cs.AI, stat.ML, cs.LG},
pdfauthor={Theodoros Ntakouris},
pdfkeywords={Neural Network, Trajectory, Compression},
}

\begin{document}
\maketitle

\begin{abstract}
	In this document, a neural network is employed in order to estimate the solution of the initial value problem in the context of non linear trajectories. Such trajectories can be subject to gravity, thrust, drag, centrifugal force, temperature, ambient air density and pressure. First, we generate a grid of trajectory points given a specified uniform density as a design parameter and then we investigate the performance of a neural network in a compression and inverse problem task: the network is trained to predict the initial conditions of the dynamics model we used in the simulation, given a target point in space. We investigate this as a regression task, with error propagation in consideration. For target points, up to a radius of 2 kilometers, the model is able to accurately predict the initial conditions of the trajectories, with sub-meter deviation. This simulation-based training process and novel real-world evaluation method is capable of computing trajectories of arbitrary dimensions.
\end{abstract}

\keywords{Neural Network \and Trajectory \and Compression \and Inverse Problem}

\section{Introduction}

The problem of estimating the initial conditions of a trajectory, given a dynamics model of a solid body of arbitrary shape or projectile, is an actively researched topic. Applications of such models include but are not limited to: probabilistic robotics and path planning, catapult systems for UAVs, guidance systems and other control and navigation problems. The broad range of systems that can benefit from such problems require robust models that produce accurate estimates both in terms of short and long distance scenarios, as well as different motion models and environment conditions (e.g. wind, temperature and pressure).

\begin{figure}
  \centering
  \subfloat[a][Error propagation]{\includegraphics[width=0.5\textwidth]{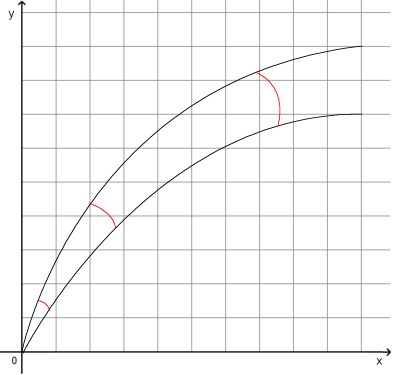} \label{fig:incerror}}
    \subfloat[b][Empirical compensation]{\includegraphics[width=0.5\textwidth]{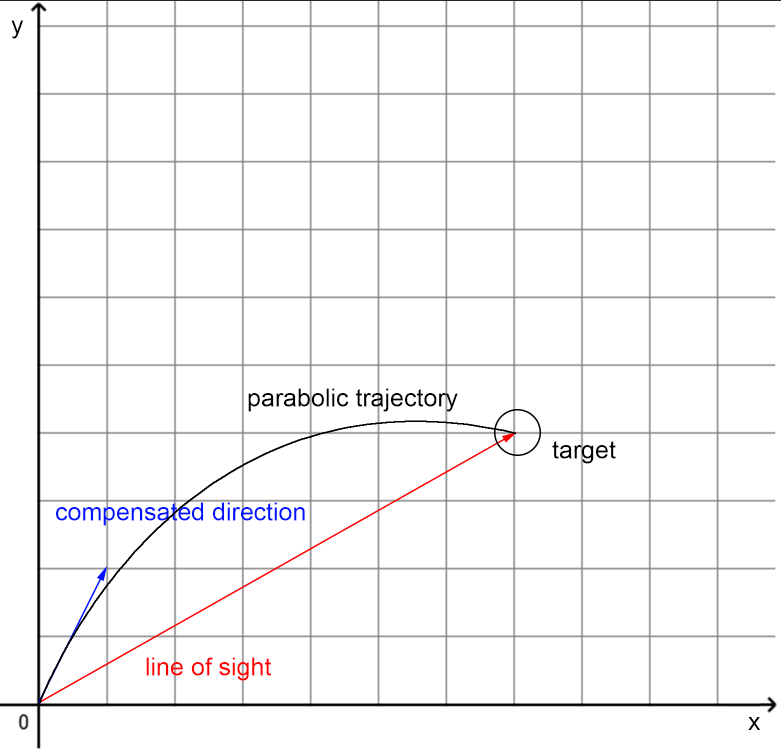}\label{fig:empcomp}}
    \caption{ }
    \label{fig:comb}
\end{figure}

During the age of antiquity, archers developed empirical methods for inclination angle and wind compensation (\autoref{fig:empcomp}). Empirical methods and simplistic mathematical models such as the Rifleman's Rule have also been applied by modern external ballistic practises. Problems that vary between ten and hundred kilometer ranges, systems without active guidance and thrusting capabilities are impossible to develop:  estimations on the initial values of the trajectory such as speed and direction are trivial to compute (\autoref{fig:incerror}). However, small errors propagate rapidly, resulting in great distance deviations in the final region of a system's effective range. Hence, models designed for smaller ranges are usually computed repeatedly, multiple times per second, predicting a system's behaviour for a few time steps ahead of time, as part of larger control systems \cite{surveyrlplanning} \cite{skydiopaper}. 

A simple and effective approach to this problem is to pre-compute a dense grid of trajectory points and perform a binary search variant for each trajectory, selecting the trajectory segment that is closer to the target point in space. Once the simulations are performed and all the data are loaded into memory, the algorithm runs within a few milliseconds. Of course, the final distance deviation is bounded by the density of the trajectory segment grid. The dynamics of some problems allow for an interpolation between the closest baked initial value combinations, to further reduce the error. Other approaches can use either simplistic non-linear trajectory path equations with closed-form approximations of the initial value problem, or complex numerical computations. Iterative methods are also included, which are robust enough to be used even in non cartesian coordinate systems, such as other geographic coordinate systems \cite{planningtrajectories}.

This report's approach generates trajectory segment data given a dense grid of initial conditions through simulation. Then, a neural network is trained to predict the initial values, given a trajectory segment. To the best of our knowledge, there are no similar neural network applications to those problems. \footnote{Specifically in the autonomous navigation and control situations, reinforcement learning algorithms are being actively developed, conditionally seeing great success purely from simulation data, applied directly to the real world, with few or no extra training data. \cite{freeman2021brax}} The solution is inspired from CocoNet \cite{coconet}, which overfits a network such that it is able to reconstruct an image: given the x and y coordinates of each pixel. By sweeping through all the grid of pixels, the network is able to reconstruct the image, while fewer parameters are stored, compared to widely adopted image compression techniques.

\section{Experiment setup}

\subsection{Problem description}
Given a target point in space $X = [x_i], \textrm{ with } x_i \in \mathbb{R^3}$, a dynamics model $D$, ambient environment conditions $E$ and projectile characteristics $P$, the magnitude and the orientation of the velocity vector is sought, which, if applied to the projectile positioned at the origin, its trajectory will orbit near the target point with a distance less than a design threshold parameter.

\subsection{Simulations}

\begin{figure}[h]
\centering
\includegraphics[width=0.3\textwidth]{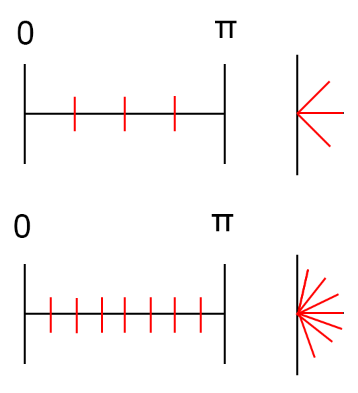}
\caption{Angular density}
\label{fig:angdens}
\end{figure}

For the experiment's dynamics model $D$, a projectile of mass $0.042 \textrm{ } kg$ is used, whose drag coefficient is equal to $0.295$, it's reference area is $0.02641 \textrm{ } m^2$ and muzzle velocity is $853 \textrm{ } m/s$ \cite{approximatedrag}. The environment ambient conditions $E$ consist of gravitational acceleration of $9.81 \textrm{ } m/s^2$ and air density of $1.225 \textrm{ } kg/m^3$. This arrangement approximates a bullet fired from a modern rifle. For the experiments, $s$ points are uniformly sampled, spanning the elevation angle linear space $ \in [0, \pi]$ (on the right half plane - see \autoref{fig:angdens}). 

By only using forces parallel to the 2D plane that the discrete angle space spans (e.g. not centrifugal and not arbitrary wind forces), the current approach generalizes in the 3D space. To achieve that, the 3D target point is mapped to a 2D one, by using the magnitude of the XY projection and the Z component of the 3D point, as 2D components. One could also re-project back to the 3D space with regards to the initial azimuth angle of the target point.

A trajectory $T$ is formulated as a list of $t_i$ points in space , indexed from $0$ up to $N$, where $N$ is the total number of points that the trajectory contains. 

\begin{equation}
\label{eqn:trajpts}
T = [t_i], i \in \mathbb{N} \textrm{ and } i \in [0, N)
\end{equation}

Starting from the origin, $s$ trajectories are simulated. The motion model $D$ is essentially a first order approximation of the projectile's movement, taking into account the initial velocity, gravity, and drag. A step time of $1e^{-4}$ is used, which is the smallest time step that executes in a reasonable time, while yielding per point errors less than $1e^{-5}$, compared to a step time of $1e^{-6}$. For the neural network training, there are no constraints on the way simulations are performed or other methods of dynamics models, such as higher-level approximations of motion models and iterative solvers. Simulations are performed for  up to $2 \textrm{ } km$ radii away from the origin. Note that each generated trajectory contains a different number of points.  

\subsection{Spatial density considerations}

Each simulation produces a list of trajectory segments. For each consecutive pair of trajectories ordered by the value of their initial elevation angle, the mean and variance of the euclidean distances between the last points (\autoref{fig:dist}) is calculated. For each experiment, the mean is much greater than the variance of those distances. Specifically, the mean is at least 3 orders of magnitude greater than the variance.

\begin{table}[!htb]
    \centering
    \begin{tabular}{||l|l||}
    \hline
        Angular Density & Euclidean Distance Mean (m) \\
        \hline\hline
         2000 &  3.14\\
         \hline
         4000 & 1.57 \\
         \hline
         6000 & 1.04 \\
         \hline
         8000 & 0.62 \\
         \hline
    \end{tabular}
    \caption{ }
    \label{tab:fp_dist}
\end{table}

As a second preprocessing step, each trajectory segment is going to be subsampled with the minimum frequency so that the maximum intra-point distance (\autoref{fig:dist}) inside the trajectories is less than the mean of the corresponding angular density. The subsampled trajectories provide uniform space coverage with a grid of points whose density is not uniform, but is bounded by this mean value.

\begin{figure}[h]
\centering
\includegraphics[width=0.5\textwidth]{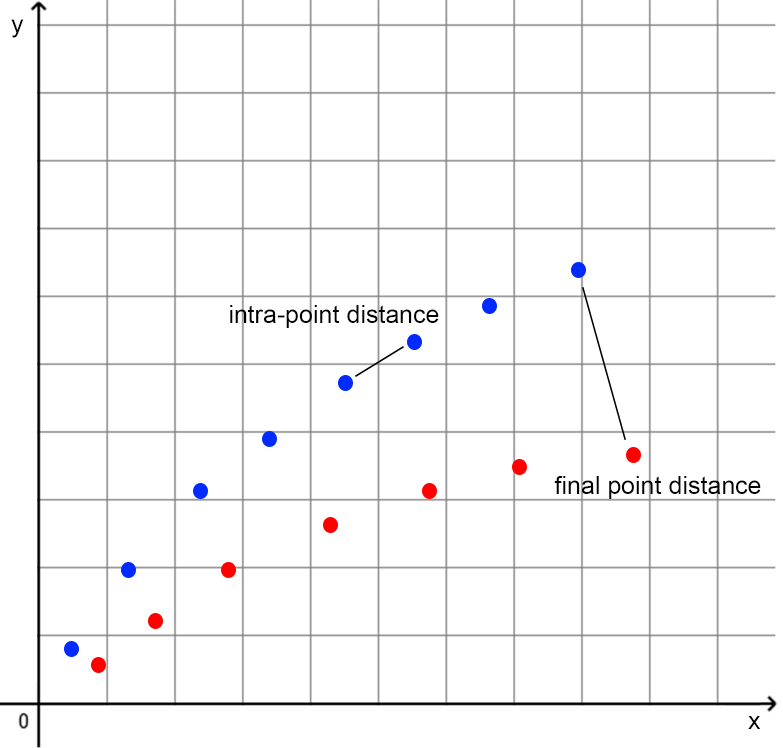}
\caption{Last point and intra-point distances}
\label{fig:dist}
\end{figure}

\subsection{Training Data}
The dataset input space consists of the following 6 features:

\begin{itemize}
    \item The X and Y coordinates of target point, normalized by the maximum simulation range. \\
    \item The norm of the target point, normalized by the maximum simulation range. \\
    \item The X and Y coordinates of the target point, normalized with regards to the norm of the vector. \\
    \item The angle between the target point and the origin, scaled between 0 and 1.
\end{itemize}

\begin{table}[htb]
    \centering
    \begin{tabular}{||l|l||}
    \hline
        Angular Density & Dataset Samples \# \\
        \hline\hline
         2000 &  1268049\\
         \hline
         4000 & 5210660 \\
         \hline
         6000 & 11722547 \\
         \hline
         8000 & 20838733 \\
         \hline
    \end{tabular}
    \caption{ }
    \label{tab:ds_samples}
\end{table}

\section{Training}
\subsection{Architecture and configuration}
\begin{figure}
\centering
\includegraphics[width=0.5\textwidth]{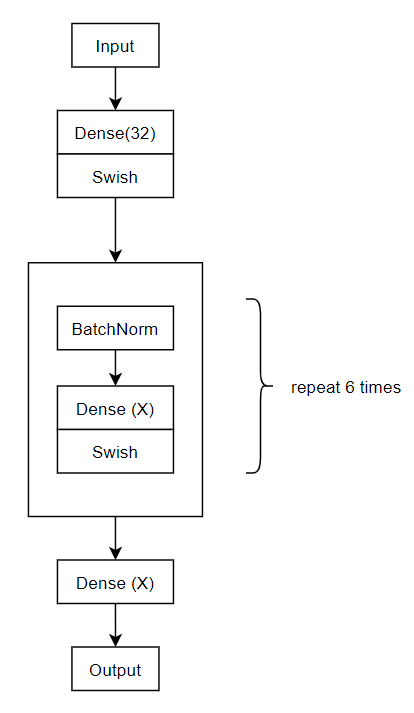}
\caption{Neural Network Architecture}
\label{fig:nnarch}
\end{figure}

A stack of fully connected layers with the swish activation function is used, independently of the task at hand. 
Out of LayerNorm \cite{layernorm} and BatchNorm \cite{batchnorm}, the latter yielded significantly better approach in a pre-norm configuration (before the fully connected layer). Experiments with dropout have no positive impact on any of the experimental runs, thus it is omitted.
The layer dimensions are as follows: $[32, 64, 128, 256, 512, 864, 1024, 2048, 4096]$. If the angular density is greater than 4096 (the size of the last fully connected layer), another layer with dimensions of 8192 is appended to the list.
Apart from the first layer, all the layer blocks (batch normalization, fully connected, dropout) are repeated 6 times (\autoref{fig:nnarch}). Between the last 2 layers, no normalization or dropout is performed. 
Mean squared error is the loss function for the regression task.

A series of experiments are performed on a TPUv3-8 \cite{tpupaper}, with a batch size of 8192. Callbacks are used for early stopping and learning rate reducing on plateaus with a minimum delta of $1e^{-2}$. Stochastic gradient descent is used as an optimizer, with a learning rate of $1e^{-2}$ and a momentum coefficient valued at $0.9$ \cite{adamsgd}. Each network is trained until convergence. 

\subsection{Results}

With the compression and lookup task, the objective for the deep neural networks would be to fully memorize the training set, similarly to an autoencoder, or CocoNet. This way, by sweeping the space (in $R^2$ or $R^3$, depending where the model is trained), one could reconstruct an approximation of the trajectory segments used to train the system, while maintaining a uniform coverage with error less than the one specified at \autoref{tab:fp_dist}.

The table below (\autoref{tab:exp_runs}) includes two means of evaluating the network's performance. Metrics such as mean squared error are sufficient for most problems, but in this case, the errors are also projected back to the granularity of the initial angular grid that is used. Different densities correspond to different pairwise distances between each angle value. Thus, for the regression task, equal loss values for different angular densities (\autoref{tab:ang_dist}) do not imply similar performance. Dense grids require less predicted angular error than sparse grids (\autoref{tab:exp_runs}). For the densities used, the distances of each consecutive angle value are as follows:

\begin{table}[!htb]
    \centering
    \begin{tabular}{||l|l|l|l||}
    \hline
        Angular Density & Pairwise Error (Uniform) \\
        \hline\hline
         2000 & $1.57e^{-3}$ \\
         \hline
         4000 & $7.851e^{-4}$ \\
         \hline
         6000 & $5.231e^{-4}$ \\
         \hline
         8000 & $3.921e^{-4}$ \\
         \hline
    \end{tabular}
    \caption{ }
    \label{tab:ang_dist}
\end{table}

For the uniform grid - density pairwise error $e_d$, "Percentage Angular Error" is defined as:

\begin{equation}
\label{eqn:pae}
        Percentage Angular Error(y_{true}, y_{pred}) = |y_{true} - y_{pred}| / e_d
\end{equation}

\begin{figure}[h]
\centering
\includegraphics[width=0.35\textwidth]{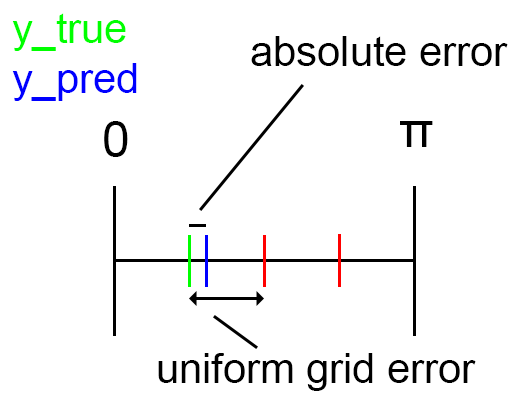}
\caption{Percentage Angular Error components}
\label{fig:dist}
\end{figure}

\begin{table}[!htb]
    \centering
    \begin{tabular}{||l|l|l|l||}
    \hline
        Angular Density & Task & MSE & Pct. Angular Error \# \\
        \hline\hline
          2000 & Regression & $4.91e^{-3}$ & 9.81 \\
         \hline
         4000 & Regression & $26e^{-3}$& 10.33 \\
         \hline
         6000 & Regression & $4.46e^{-4}$& 26.7 \\
         \hline
         8000 & Regression & $3.29e^{-4}$ & 36.7 \\
         \hline
    \end{tabular}
    \caption{ }
    \label{tab:exp_runs}
\end{table}

The percentage of angular error has less than 50\% value, which signifies that the network has indeed learned to classify target points effectively. By quantizing this continuous output angle value back to the discrete angle linear space, one is able to precisely infer the initial orientation at the origin, which led to each trajectory segment.

\section{Conclusions}

All in all, not all problems are suitable for empirical machine learning methods such as neural network training. For systems with error tolerance similar to the performed experiments - which is a few meters at most, the non-linear nature of neural networks and deep learning can be used to perform segmentation of this inverse problem of target points in 2D and 3D space. A simple dataset construction scheme is presented, accompanied by a training and evaluation framework which is easy to extend in other dimensions, ambient conditions, dynamics models, and forces.

\bibliographystyle{unsrtnat}
\bibliography{references}  

\begin{thebibliography}{10}
\providecommand{\natexlab}[1]{#1}
\providecommand{\url}[1]{\texttt{#1}}
\expandafter\ifx\csname urlstyle\endcsname\relax
  \providecommand{\doi}[1]{doi: #1}\else
  \providecommand{\doi}{doi: \begingroup \urlstyle{rm}\Url}\fi

\bibitem[Aradi(2020)]{surveyrlplanning}
Szilárd Aradi.
\newblock Survey of deep reinforcement learning for motion planning of
  autonomous vehicles, 2020.

\bibitem[Bonatti et~al.(2019)Bonatti, Ho, Wang, Choudhury, and
  Scherer]{skydiopaper}
Rogerio Bonatti, Cherie Ho, Wenshan Wang, Sanjiban Choudhury, and Sebastian
  Scherer.
\newblock Towards a robust aerial cinematography platform: Localizing and
  tracking moving targets in unstructured environments, 2019.

\bibitem[Ezair et~al.(2014)Ezair, Tassa, and Shiller]{planningtrajectories}
Ben Ezair, Tamir Tassa, and Zvi Shiller.
\newblock Planning high order trajectories with general initial and final
  conditions and asymmetric bounds.
\newblock \emph{The International Journal of Robotics Research}, 33\penalty0
  (6):\penalty0 898--916, 2014.
\newblock \doi{10.1177/0278364913517148}.
\newblock URL \url{https://doi.org/10.1177/0278364913517148}.

\bibitem[Freeman et~al.(2021)Freeman, Frey, Raichuk, Girgin, Mordatch, and
  Bachem]{freeman2021brax}
C.~Daniel Freeman, Erik Frey, Anton Raichuk, Sertan Girgin, Igor Mordatch, and
  Olivier Bachem.
\newblock Brax -- a differentiable physics engine for large scale rigid body
  simulation, 2021.

\bibitem[Bricman and Ionescu(2018)]{coconet}
Paul~Andrei Bricman and Radu~Tudor Ionescu.
\newblock Coconet: A deep neural network for mapping pixel coordinates to color
  values, 2018.

\bibitem[de~Carpentier(2014)]{approximatedrag}
Giliam J.~P. de~Carpentier.
\newblock Analytical ballistic trajectories with approximately linear drag.
\newblock \emph{Int. J. Comput. Games Technol.}, 2014, January 2014.
\newblock ISSN 1687-7047.
\newblock \doi{10.1155/2014/463489}.
\newblock URL \url{https://doi.org/10.1155/2014/463489}.

\bibitem[Ba et~al.(2016)Ba, Kiros, and Hinton]{layernorm}
Jimmy~Lei Ba, Jamie~Ryan Kiros, and Geoffrey~E. Hinton.
\newblock Layer normalization, 2016.

\bibitem[Ioffe and Szegedy(2015)]{batchnorm}
Sergey Ioffe and Christian Szegedy.
\newblock Batch normalization: Accelerating deep network training by reducing
  internal covariate shift, 2015.

\bibitem[Jouppi et~al.(2020)Jouppi, Yoon, Kurian, Li, Patil, Laudon, Young, and
  Patterson]{tpupaper}
Norman~P. Jouppi, Doe~Hyun Yoon, George Kurian, Sheng Li, Nishant Patil, James
  Laudon, Cliff Young, and David Patterson.
\newblock A domain-specific supercomputer for training deep neural networks.
\newblock \emph{Commun. ACM}, 63\penalty0 (7):\penalty0 67–78, June 2020.
\newblock ISSN 0001-0782.
\newblock \doi{10.1145/3360307}.
\newblock URL \url{https://doi.org/10.1145/3360307}.

\bibitem[Keskar and Socher(2017)]{adamsgd}
Nitish~Shirish Keskar and Richard Socher.
\newblock Improving generalization performance by switching from adam to sgd,
  2017.

\end{thebibliography}
\end{document}